%% file: main.tex
\documentclass[letterpaper]{article} %
\usepackage[preprint]{aaai2027}  %
\usepackage[hyphens]{url}  %
\usepackage{graphicx} %
\usepackage{natbib}  %
\usepackage{caption} %
\usepackage{booktabs}
\usepackage{colortbl}
\usepackage{amsmath}

\title{BrainTaskonomy: Learning How to Pretrain and What to Transfer in fMRI Foundation Models}
\input{authors.tex}

\begin{document}
\setlength{\footskip}{18pt}
\pagestyle{plain}
\maketitle
\thispagestyle{plain}

\begin{abstract}
fMRI foundation models increasingly aggregate heterogeneous data across
brain states, cohorts, and acquisition settings, yet pretraining domains
are commonly treated as a flat mixture and downstream tasks are adapted
independently. We study whether measured learning relations can organize
both stages without modifying the backbone. During pretraining, a
lightweight Brain-DiT proxy estimates difficulty and directed facilitation
across ten fMRI domains, yielding a priority-guided cumulative domain
curriculum combined with high-to-low-noise timestep scheduling and joint
consolidation. During adaptation, controlled first- and higher-order
transfer across fifteen tasks constructs a directed taskonomy, from which
budgeted integer programming (BIP) selects directly supervised source
tasks and target-specific routes. The joint priority-domain and high-to-low-timestep curriculum reduces v-NMSE, PSD-NMSE, and FC-MSE by \(6.5\%\), \(16.3\%\), and
\(10.5\%\), respectively, relative to uniform sampling over both dimensions, and shows strong downstream performance
across six in- and out-of-domain tasks. The taskonomy
reveals asymmetric, target-dependent transfer, while exploratory
sealed-test evaluation shows larger descriptive gains for BIP policies
when higher-order route spaces are available than for matched random
controls. Together, these findings support organizing fMRI pretraining
and adaptation by measured learning relations rather than treating
domains and tasks as independent flat sets.
\end{abstract}

\section{Introduction}

Brain foundation models have emerged across EEG and fMRI, using
large-scale self-supervised pretraining on heterogeneous neural
recordings to learn reusable representations that are subsequently
adapted with task-specific supervision
\cite{guo2026brain,wang2024eegpt,jiang2024large,xia2026brain,xia2026brainworld,wang2026omni,wang2026towards,dong2024brain}. As their pretraining corpora and
downstream task spaces expand, two allocation questions become
increasingly important: which training experiences should be prioritized
to establish transferable representations, and which downstream
capabilities should receive direct supervision so that others can be
acquired through transfer?

\begin{figure}[t]
    \centering
    \includegraphics[width=\columnwidth]{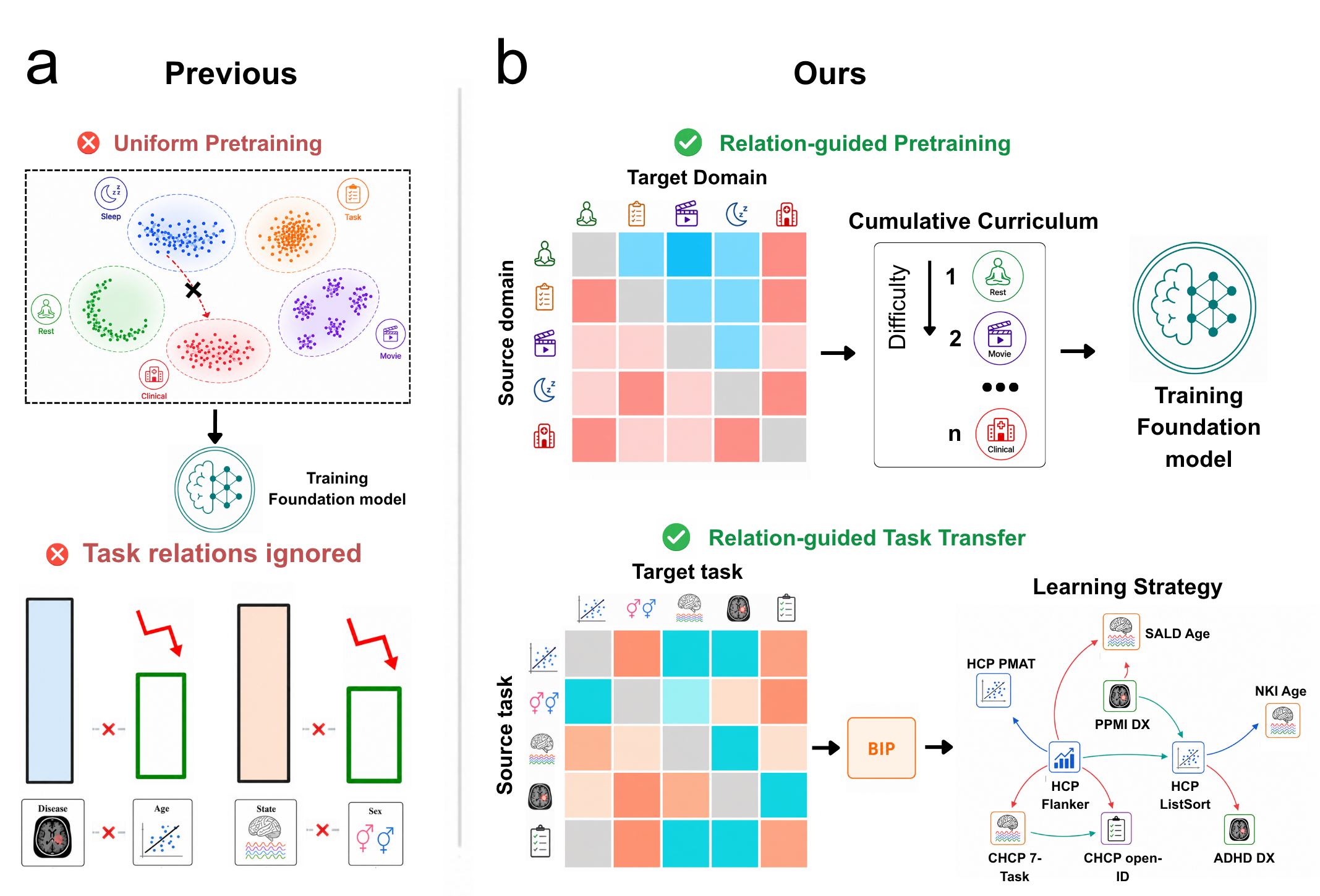}
    \caption{Motivation. (a) Existing pipelines mix heterogeneous pretraining domains and adapt downstream tasks independently, ignoring their relations. (b) We use directed domain relations to construct a pretraining curriculum and downstream task relations to jointly plan supervision and transfer across targets.}
    \label{fig:braincompass-overview}
\end{figure}

\begin{figure*}[!t]
    \centering
    \includegraphics[width=\textwidth]{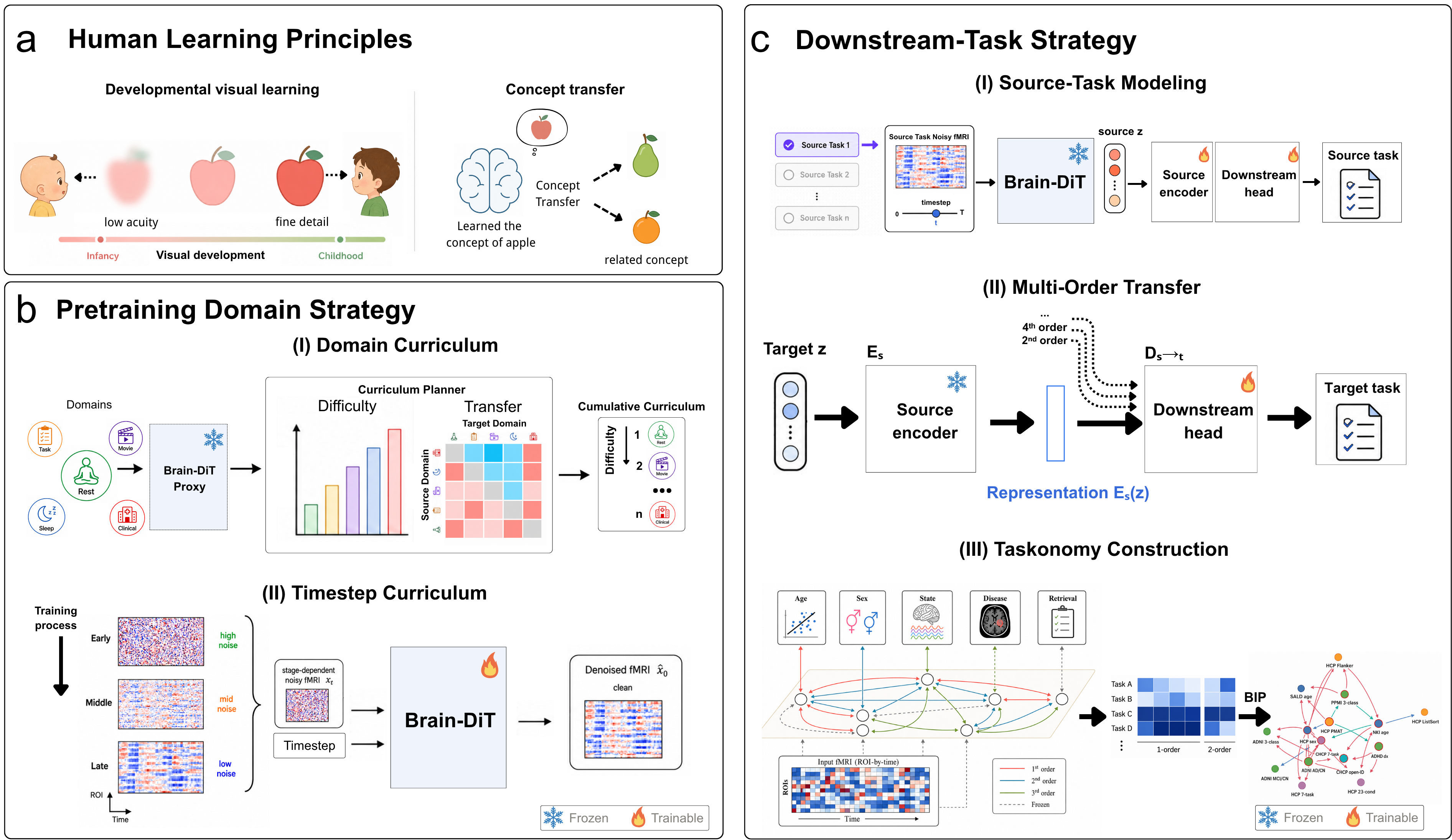}
    \caption{Overview of the proposed framework.
    (a) Human-inspired motivation: visual development proceeds from coarse to fine, while learned concepts facilitate related learning.
    (b) Two-level pretraining curriculum: domain difficulty and directed facilitation define cumulative domain stages, while a difficulty-motivated high-to-low-noise schedule progressively expands the active timestep range.
    (c) Downstream taskonomy: source-task encoders \(E_s\) are trained on
    fixed Brain-DiT features and frozen during taskonomy construction.
    For a source set \(S\), the source bottlenecks are concatenated and
    passed to a target-specific readout \(d_{S\rightarrow t}\), trained on
    low-shot target data to estimate first- and higher-order transfer. BIP then selects
    supervised source tasks and transfer paths under a limited budget.}
    \label{fig:braincompass-framework}
\end{figure*}

These questions are particularly salient for fMRI foundation models. To
support diverse cognitive and clinical applications, recent models
increasingly aggregate resting-state, task-evoked, naturalistic,
lifespan, and clinical data
\cite{xia2026brain,xia2026brainworld}. Yet these domains are commonly
mixed randomly or in proportion to dataset size from the beginning of
pretraining, despite substantial differences in neural dynamics,
populations, acquisition protocols, signal quality, and scale. Some
domains may be readily learned and provide transferable foundations for
others, whereas specialized or long-tailed domains may be more difficult
or overshadowed by larger datasets. Meanwhile, downstream tasks such as
demographic prediction, cognitive assessment, clinical diagnosis,
brain-state decoding, and individual identification are usually adapted
independently, despite scarce labels and potentially shared neural
representations. fMRI foundation models therefore face two related
allocation problems: how to distribute pretraining computation across
heterogeneous domains, and how to distribute limited supervision across
interdependent downstream tasks.

Several lines of research provide complementary guidance for addressing
these problems. Human visual experience develops from coarse,
low-acuity observations toward increasingly fine-grained inputs, and
training artificial vision models with a corresponding developmental
visual diet produces more robust representations than randomly mixing
the same experiences
\cite{braddick2011development,brown2015contrast,
lu2026developmentalVisualDiet}. Curriculum learning and data-mixture
optimization further show that examples, domains, and skills need not be
treated as exchangeable training units
\cite{bengio2009curriculum,chen2023skill,xie2023doremi}.
In diffusion models, denoising objectives at different timesteps exhibit
different optimization difficulty, motivating progressive schedules from
high-noise recovery to low-noise refinement
\cite{kim2025denoising}. Complementarily, Taskonomy and its fMRI
extensions reveal directed first- and higher-order transfer relations and
use them to allocate supervision under limited budgets
\cite{zamir2018taskonomy,qu2024uncovering,xia2026beyond}.
Together, these studies suggest a common principle: learning priority
should depend not only on intrinsic learnability, but also on how strongly
an object supports subsequent learning.

Based on this principle, we study learning relations across pretraining
domains and downstream tasks. Our contributions are threefold:
\begin{itemize}
    \item \textbf{Learning relations.}
    We characterize fixed-budget difficulty and directed facilitation
    across ten fMRI domains, together with first- and higher-order
    transfer across fifteen downstream tasks.

    \item \textbf{Learning how to pretrain.}
    We use domain relations to construct a cumulative domain curriculum,
    combined with high-to-low-noise timestep scheduling and joint
    consolidation.

    \item \textbf{Learning what to transfer.}
    We use the downstream taskonomy and BIP to select supervised source
    tasks and target-specific transfer routes under limited budgets.
\end{itemize}

Experiments demonstrate improved pretraining fidelity and in-domain
transfer, while exploratory sealed-test results suggest benefits from
taskonomy-guided policies with access to higher-order route spaces.

\FloatBarrier

\begin{figure}[!t]
    \centering
    \includegraphics[width=\columnwidth]{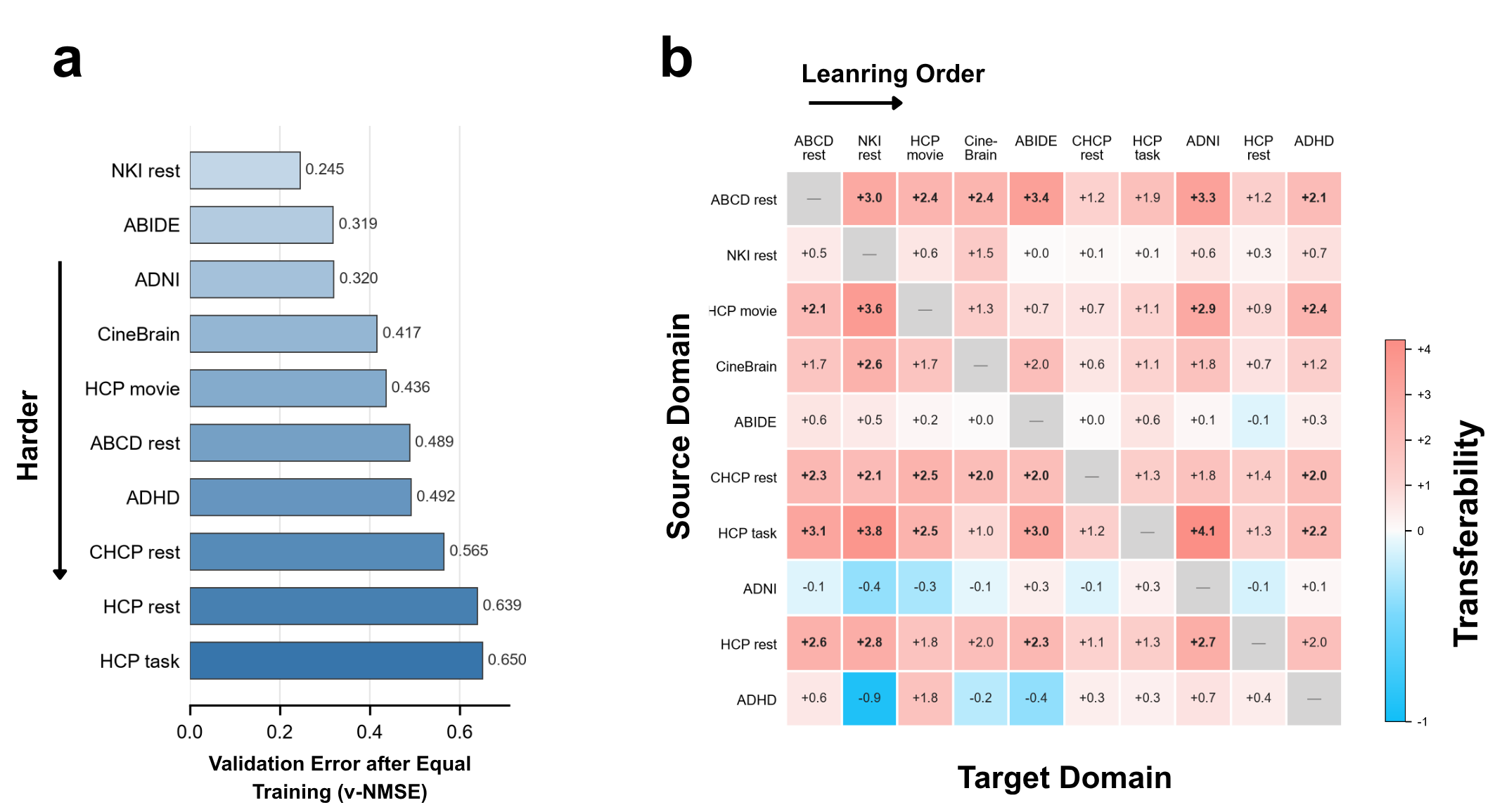}
    \caption{Domain difficulty and directed facilitation.
    (a) Fixed-budget domain difficulty measured by v-NMSE.
    (b) Directed facilitation between source and target domains.
    Red denotes facilitation and blue denotes interference.
    Difficulty and facilitation jointly determine the curriculum order.}
    \label{fig:domain-structure-curriculum}
\end{figure}

\section{Related Work}

\subsection{fMRI Foundation Models}

fMRI foundation models have progressed through richer objectives and finer spatial scales. At the ROI level, BrainLM reconstructs masked signals, Brain-JEPA predicts latent targets, and BrainMass combines functional-connectivity reconstruction with representation alignment~\cite{ortega2024brainlm,dong2024brain,yang2024brainmass}. Voxel-level models learn 4D dynamics through efficient temporal modeling or dynamic patching~\cite{wang2026towards,wang2026omni,wang2025slim}. Brain-DiT spans resting, task, naturalistic, disease, and sleep states, while BrainWorld extends diffusion pretraining to voxel-level whole-brain dynamics~\cite{xia2026brain,xia2026brainworld}. Yet these domains remain mixed using fixed or random schedules, leaving their learning structure unexplored.

\subsection{Curriculum Learning and Transfer-Aware Resource Allocation}

Foundation models depend strongly on how heterogeneous data are sampled and ordered. DoReMi, DoGE, and RegMix optimize domain proportions using proxy or cross-domain signals~\cite{xie2023doremi,fan2024doge,liu2025regmix}. Skill-It organizes skills through prerequisite relations~\cite{chen2023skill}, while diffusion curricula progressively schedule timesteps with different convergence difficulty~\cite{kim2025denoising}. Taskonomy instead estimates directed first- and higher-order transfer and selects supervised source tasks under a limited budget~\cite{zamir2018taskonomy,achille2019task2vec,standley2020tasks,fifty2021efficiently}. Existing work therefore treats mixture proportions, learning order, objective difficulty, and transfer utility separately; these principles remain unintegrated for multi-state fMRI pretraining and downstream supervision.

\subsection{fMRI Heterogeneity and Transfer Structure}

fMRI heterogeneity spans brain states, populations, diagnoses, acquisition protocols, resolutions, and preprocessing spaces. FlexiBrain accommodates this variability through native-space dynamic patching and resolution-adaptive embeddings, and reveals asymmetric transfer across spaces and diagnostic populations~\cite{wang2026flexibrain}. Cognitive taskonomies similarly show that fMRI task states are interdependent~\cite{qu2024uncovering}. Single-source transfer is directed, while multi-source transfer depends on source composition and cannot be inferred from pairwise relations alone~\cite{xia2026beyond}. Thus, fMRI heterogeneity is not only a nuisance, but also transferable structure that can guide learning.

\begin{table}[!t]
    \centering
    \setlength{\tabcolsep}{4.5pt}
    \renewcommand{\arraystretch}{1.03}
    \resizebox{\columnwidth}{!}{%
    \begin{tabular}{lccc}
        \toprule
        \multicolumn{4}{c}{\textbf{(a) Uniform timestep sampling}} \\
        \midrule
        \textbf{Domain Policy}
        & \textbf{v-NMSE} $\downarrow$
        & \textbf{PSD-NMSE} $\downarrow$
        & \textbf{FC-MSE} $\downarrow$ \\
        \midrule
        Uniform & 0.4148 & 0.1506 & 0.0611 \\
        Random & 0.3950 & 0.1377 & 0.0565 \\
        Priority & 0.3936 & 0.1349 & 0.0567 \\
        \midrule
        \multicolumn{4}{c}{\textbf{(b) High-to-low timestep curriculum}} \\
        \midrule
        \textbf{Domain Policy}
        & \textbf{v-NMSE} $\downarrow$
        & \textbf{PSD-NMSE} $\downarrow$
        & \textbf{FC-MSE} $\downarrow$ \\
        \midrule
        Uniform & 0.3885 & 0.1302 & 0.0548 \\
        Random & 0.3893 & 0.1288 & 0.0548 \\
        Priority & \textbf{0.3880} & \textbf{0.1261} & \textbf{0.0547} \\
        \bottomrule
    \end{tabular}
    }
    \caption{Pretraining ablations defined jointly by timestep policy (panels) and domain policy (rows). Uniform, Random, and Priority denote uniform mixing, random ordering, and priority-guided ordering. Training-step counts and further details are provided in the supplementary material. Lower errors are better; bold is best.}
    \label{tab:curriculum-reconstruction}
\end{table}

\section{Method}
\label{sec:method}

\begin{table*}[!t]
    \centering
    \begingroup
    \setlength{\tabcolsep}{2.2pt}
    \renewcommand{\arraystretch}{1.02}

    \resizebox{\textwidth}{!}{%
    \begin{tabular}{l*{12}{c}}
        \toprule
        \textbf{Method}
        & \multicolumn{8}{c}{\textbf{In-domain}}
        & \multicolumn{4}{c}{\textbf{Out-of-domain}} \\
        \cmidrule(lr){2-9}
        \cmidrule(lr){10-13}

        & \multicolumn{2}{c}{\textbf{ABIDE}}
        & \multicolumn{2}{c}{\textbf{NKI}}
        & \multicolumn{2}{c}{\textbf{ABCD}}
        & \multicolumn{2}{c}{\textbf{HCP}}
        & \multicolumn{2}{c}{\textbf{SALD}}
        & \multicolumn{2}{c}{\textbf{PPMI}} \\
        \cmidrule(lr){2-3}
        \cmidrule(lr){4-5}
        \cmidrule(lr){6-7}
        \cmidrule(lr){8-9}
        \cmidrule(lr){10-11}
        \cmidrule(lr){12-13}

        & MSE $\downarrow$ & $r$ $\uparrow$
        & MSE $\downarrow$ & $r$ $\uparrow$
        & ACC $\uparrow$ & F1 $\uparrow$
        & ACC $\uparrow$ & F1 $\uparrow$
        & MSE $\downarrow$ & $r$ $\uparrow$
        & ACC $\uparrow$ & F1 $\uparrow$ \\

        \midrule
        \multicolumn{13}{l}{\textbf{Baselines}} \\

        BrainLM
        & .91$\pm$.01 & .24$\pm$.02
        & .50$\pm$.02 & .68$\pm$.01
        & 59.24$\pm$1.03 & 58.74$\pm$.70
        & 62.71$\pm$4.43 & 61.74$\pm$4.72
        & .68$\pm$.06 & .62$\pm$.06
        & \underline{68.79$\pm$3.25} & 55.46$\pm$1.57 \\

        Brain-JEPA
        & .98$\pm$.07 & .17$\pm$.02
        & 1.03$\pm$.08 & .33$\pm$.03
        & -- & --
        & 69.97$\pm$2.73 & 69.17$\pm$3.55
        & 1.13$\pm$.11 & .30$\pm$.06
        & 63.12$\pm$5.35 & 53.85$\pm$2.90 \\

        BrainMass
        & .70$\pm$.04 & .50$\pm$.04
        & .60$\pm$.06 & .62$\pm$.04
        & 57.92$\pm$1.19 & 57.86$\pm$1.18
        & 67.65$\pm$1.02 & 66.93$\pm$1.78
        & .70$\pm$.08 & .63$\pm$.06
        & 63.12$\pm$4.43 & 49.47$\pm$1.16 \\

        \midrule
        \multicolumn{13}{l}{\textbf{Uniform timestep sampling}} \\

        Uniform
        & .53$\pm$.01 & .63$\pm$.00
        & \underline{.30$\pm$.02} & \underline{.86$\pm$.01}
        & 56.84$\pm$.84 & 56.36$\pm$.95
        & 79.70$\pm$5.45 & 79.39$\pm$5.77
        & .501$\pm$.001 & .726$\pm$.002
        & 63.83$\pm$9.75 & 52.14$\pm$2.67 \\

        Random
        & .53$\pm$.05 & .63$\pm$.06
        & .31$\pm$.02 & .84$\pm$.01
        & \underline{66.19$\pm$.11} & \underline{66.10$\pm$.10}
        & 79.21$\pm$4.40 & 79.01$\pm$4.37
        & .538$\pm$.014 & .655$\pm$.004
        & 68.09$\pm$2.13
        & \underline{60.39$\pm$.94} \\

        Priority
        & .51$\pm$.04 & .65$\pm$.03
        & .31$\pm$.02 & .85$\pm$.01
        & 63.79$\pm$5.04
        & 63.48$\pm$5.51
        & 81.02$\pm$3.48 & 80.99$\pm$3.52
        & .466$\pm$.014 & \underline{.754$\pm$.002}
        & 67.38$\pm$4.91 & 53.20$\pm$2.03 \\

        \midrule
        \multicolumn{13}{l}{\textbf{High-to-low timestep curriculum}} \\

        Uniform
        & .46$\pm$.04 & \underline{.69$\pm$.03}
        & .32$\pm$.03 & .84$\pm$.02
        & 61.59$\pm$2.01 & 61.16$\pm$2.01
        & \underline{82.51$\pm$3.37}
        & \underline{82.44$\pm$3.34}
        & \underline{.455$\pm$.019}
        & \textbf{.756$\pm$.002}\textsuperscript{*}
        & 65.96$\pm$3.69
        & 56.36$\pm$2.59 \\

        Random
        & \underline{.45$\pm$.02} & \underline{.69$\pm$.01}
        & .31$\pm$.01 & .85$\pm$.01
        & 63.14$\pm$.82 & 62.96$\pm$.65
        & 81.19$\pm$1.78 & 80.80$\pm$2.41
        & \textbf{.406$\pm$.006}\textsuperscript{*} & .729$\pm$.005
        & 68.09$\pm$2.13
        & 58.97$\pm$.45 \\

        \rowcolor{gray!15}
        Priority
        & \textbf{.43$\pm$.02}\textsuperscript{*}
        & \textbf{.71$\pm$.01}\textsuperscript{*}
        & \textbf{.27$\pm$.01}\textsuperscript{*}
        & \textbf{.87$\pm$.01}\textsuperscript{*}
        & \textbf{66.38$\pm$.32}\textsuperscript{*}
        & \textbf{66.26$\pm$.45}\textsuperscript{*}
        & \textbf{84.32$\pm$3.72}\textsuperscript{*}
        & \textbf{84.29$\pm$3.73}\textsuperscript{*}
        & .458$\pm$.004
        & .746$\pm$.004
        & \textbf{71.63$\pm$3.25}\textsuperscript{*}
        & \textbf{67.42$\pm$.78}\textsuperscript{*} \\

        \bottomrule
    \end{tabular}
    }

    \endgroup

    \caption{
    Downstream comparison of curriculum-pretrained Brain-DiT variants
    and existing fMRI foundation models under the Omni-fMRI
    benchmark~\cite{wang2026omni}.
    Brain-DiT variants are organized by timestep-policy panels, following
    Table~\ref{tab:curriculum-reconstruction}. Within each panel, Uniform,
    Random, and Priority in the Method column are domain policies denoting
    uniform mixing, random ordering, and priority-guided ordering,
    respectively.
    Results are reported as mean$\pm$standard deviation; ``--'' denotes
    unavailable results.
    Bold and underlined values indicate the best and second-best results,
    respectively.
    Gray highlights Priority under the high-to-low timestep
    curriculum.
    \textsuperscript{*} indicates a large effect relative to the strongest
    external baseline (Cohen's $d\geq0.8$).
    }

    \label{tab:downstream-comparison}
\end{table*}

Without modifying the Brain-DiT backbone, we study how to organize
heterogeneous domains and diffusion timesteps during pretraining, and how
to allocate limited supervision across downstream tasks. Given a clean
fMRI window $\mathbf{x}_0$, timestep $t$, and Gaussian noise
$\boldsymbol{\epsilon}\sim\mathcal{N}(\mathbf{0},\mathbf{I})$, the noisy
input and velocity target are
\begin{equation}
\begin{aligned}
\mathbf{x}_t
&=
\sqrt{\bar{\alpha}_t}\mathbf{x}_0
+
\sqrt{1-\bar{\alpha}_t}\boldsymbol{\epsilon},\\
\mathbf{v}_t
&=
\sqrt{\bar{\alpha}_t}\boldsymbol{\epsilon}
-
\sqrt{1-\bar{\alpha}_t}\mathbf{x}_0.
\end{aligned}
\label{eq:diffusion-velocity}
\end{equation}
We first estimate domain difficulty and directed facilitation to construct
a nested domain--timestep curriculum, and then model downstream task
transfer to derive budget-aware supervision policies.

\subsection{Structure-Guided Curriculum Pretraining}
\label{sec:curriculum-method}

\paragraph{Domain difficulty.}
A domain is defined as a dataset--state pair with a distinct cohort and
acquisition distribution.
Let \(\{\mathcal{D}_i\}_{i=1}^{M}\) denote the candidate fMRI domains.
We train the same lightweight Brain-DiT proxy independently on each domain
using identical architectures, optimization budgets, and uniform
timestep sampling. Let
\(\mathcal{B}_i=
\{(\mathbf{x}_0^{(n)},\boldsymbol{\epsilon}^{(n)})\}_{n=1}^{N_i}\)
be a frozen validation bank for domain \(i\). Its normalized
velocity-prediction error at timestep \(t\) is
\begin{equation}
\ell_i(t)=
\frac{
\sum_{n=1}^{N_i}
\left\|
\mathbf{v}_t^{(n)}
-
\mathbf{v}_{\theta_i}
\left(\mathbf{x}_t^{(n)},t\right)
\right\|_F^2
}{
\sum_{n=1}^{N_i}
\left\|\mathbf{v}_t^{(n)}\right\|_F^2
+\varepsilon
},
\label{eq:domain-nmse}
\end{equation}
where the denominator normalizes the residual by the energy of the
ground-truth velocity target. We define domain difficulty by averaging
over a fixed timestep grid \(\mathcal{G}\):
\begin{equation}
D_i=
\frac{1}{|\mathcal{G}|}
\sum_{t\in\mathcal{G}}\ell_i(t).
\label{eq:domain-difficulty}
\end{equation}
A lower \(D_i\) indicates that domain \(i\) is easier for the proxy model
to learn.

\paragraph{Compute-matched directed facilitation and priority.}
To quantify whether prioritizing \(i\) facilitates subsequent learning of
\(j\), we compare compute-matched policies. The source policy allocates
the entire source-stage budget to \(i\), whereas the reference distributes
it over a domain-balanced mixture excluding \(j\). Starting from the same
initialization, both then continue on the same target bank from \(j\),
with matched windows, timesteps, noise, and optimization budgets.
Let \(L_{i\rightarrow j}\) and
\(L_{\mathrm{ref}\rightarrow j}\) denote their final target NMSEs. We
define
\begin{equation}
F(i\rightarrow j)=
\frac{
L_{\mathrm{ref}\rightarrow j}
-
L_{i\rightarrow j}
}{
L_{\mathrm{ref}\rightarrow j}
}.
\label{eq:domain-facilitation}
\end{equation}
A positive value means prioritizing \(i\) better prepares the model for
\(j\) than generic non-target pretraining; a negative value denotes a
relative disadvantage. Thus, \(F(i\rightarrow j)\) measures
policy-relative directed facilitation, not the leave-one-domain-out
marginal contribution of \(i\).

We summarize each domain's global transfer value by its mean outgoing
facilitation and combine it with difficulty:
\begin{equation}
\bar{F}_i=
\frac{1}{M-1}\sum_{j\neq i}F(i\rightarrow j),
\qquad
p_i=z(D_i)-z(\bar{F}_i).
\label{eq:domain-priority}
\end{equation}
Domains are introduced in ascending order of \(p_i\), prioritizing those
that are easier to learn and more beneficial to subsequent domains. The
order is fixed before full-scale pretraining.

\paragraph{SNR-partitioned timestep curriculum.}
Following difficulty-based diffusion curriculum learning
\cite{kim2025denoising}, we progressively expand denoising from high-noise
to low-noise timesteps. For each timestep
\(t\in\{1,\ldots,T_{\mathrm{diff}}\}\), we compute
\begin{equation}
r_t=
\log\frac{\bar{\alpha}_t}{1-\bar{\alpha}_t}.
\label{eq:log-snr}
\end{equation}
The ordered timestep sequence is divided into \(K_{\tau}\) contiguous
clusters using boundaries
\(0=b_0<b_1<\cdots<b_{K_{\tau}}=T_{\mathrm{diff}}\), with
\[
C_k=\{b_{k-1}+1,\ldots,b_k\}.
\]
The boundaries are obtained once by constrained dynamic programming:
\begin{equation}
\begin{aligned}
\min_{\{b_k\}}\quad &
\sum_{k=1}^{K_{\tau}}
\sum_{t\in C_k}
\left|
r_t-\operatorname{median}_{u\in C_k}r_u
\right|,\\
\text{s.t.}\quad &
n_{\min}\leq |C_k|\leq n_{\max}.
\end{aligned}
\label{eq:snr-clustering}
\end{equation}
We index the clusters such that \(C_1\) contains the lowest-noise
timesteps with the highest log-SNR, whereas \(C_{K_{\tau}}\) contains the
highest-noise timesteps with the lowest log-SNR.

\paragraph{Nested curriculum and plateau gates.}
Let \(\sigma=(\sigma_1,\ldots,\sigma_M)\) denote the priority-guided
domain order. Domains are introduced cumulatively under the initial
high-noise support \(C_{a_0:K_\tau}\), with all previously activated
domains retained through replay. A stage advances when the relevant
fixed-bank loss fails to improve by at least \(\delta\) from its
stage-best value for a prescribed patience: \(K_d\) evaluations on the
newest domain trigger the next domain, while, after all domains are
active, \(K_t\) evaluations on the domain-balanced loss trigger the next
lower-noise cluster,
\begin{equation}
C_{a:K_\tau}\rightarrow C_{a-1:K_\tau}.
\label{eq:timestep-unlock}
\end{equation}
Previously activated domains and timestep clusters remain available
throughout training. Once all clusters are active, the model undergoes
uniform joint consolidation over all domains and timesteps.

\subsection{Downstream Taskonomy}
\label{sec:downstream-taskonomy}

\paragraph{First-order transfer.}
Given a fixed Brain-DiT feature extractor
\(\Phi_{\mathrm{BD}}\), each scan is represented as
\(z=\Phi_{\mathrm{BD}}(x)\). For each source task \(s\), we train a
standardized encoder \(E_s\) and head \(h_s\). To evaluate
\(s\rightarrow t\), we discard \(h_s\), freeze \(E_s\), and train a new
target head on a fixed low-shot support set. We compare this route with a
matched self-transfer reference \(t\rightarrow t\), whose encoder is
trained on task \(t\) and matched in architecture, target support, head
capacity, initialization, and optimization budget. With all metrics
oriented such that larger values are better, the first-order affinity is
\begin{equation}
G_1(s\rightarrow t)=
M_t(s\rightarrow t)-M_t(t\rightarrow t),
\qquad s\neq t.
\label{eq:first-order-gain}
\end{equation}

\paragraph{Higher-order transfer.}
For a source set \(S\) of order \(m=|S|>1\), all source encoders remain
frozen. Each source representation is projected through a source-fitted
PCA bottleneck, after which the projected representations are
concatenated and passed to a target-specific ridge readout:
\begin{equation}
r_S(z)=
\operatorname{Concat}_{s\in S}
\left[P_s\!\left(E_s(z)\right)\right],
\qquad
\widehat y_t=d_{S\rightarrow t}\!\left(r_S(z)\right).
\label{eq:higher-order-transfer}
\end{equation}
Here, \(P_s\) denotes the PCA projection fitted from source-task
representations. Only the target readout \(d_{S\rightarrow t}\) is
fitted during Taskonomy construction. Each route is compared with
an order-matched self-transfer reference
\(t^{\times m}\rightarrow t\), which uses \(m\) independently trained
target-task encoders and matches the transfer route in fusion/head
capacity, target support, initialization, and optimization budget:
\begin{equation}
G_m(S\rightarrow t)=
M_t(S\rightarrow t)-M_t(t^{\times m}\rightarrow t).
\label{eq:higher-order-gain}
\end{equation}

Because downstream tasks use heterogeneous metrics, we calibrate each
gain using a prespecified target-specific increment,
\(\widetilde G_m(S\rightarrow t)=G_m(S\rightarrow t)/\delta_t\).
These calibrated affinities define the directed downstream taskonomy.
Target-wise \(z\)-scores are used only for visualization, whereas route
selection uses \(\widetilde G_m\) directly.

\subsection{Budget-Aware Transfer Planning}
\label{sec:bip}

BIP uses only construction-validation taskonomy results. Let
\(x_s\in\{0,1\}\) indicate whether source task \(s\) receives direct
supervision, and \(y_{t,r}\in\{0,1\}\) indicate whether route \(r\) is
assigned to target \(t\). Each route requires a source set \(S_r\) and has
utility
\begin{equation}
u_{t,r}=
\widetilde{G}^{\mathrm{cv}}_{|S_r|}(r\rightarrow t).
\label{eq:bip-utility}
\end{equation}

\begin{figure}[!t]
    \centering
    \includegraphics[width=\columnwidth]{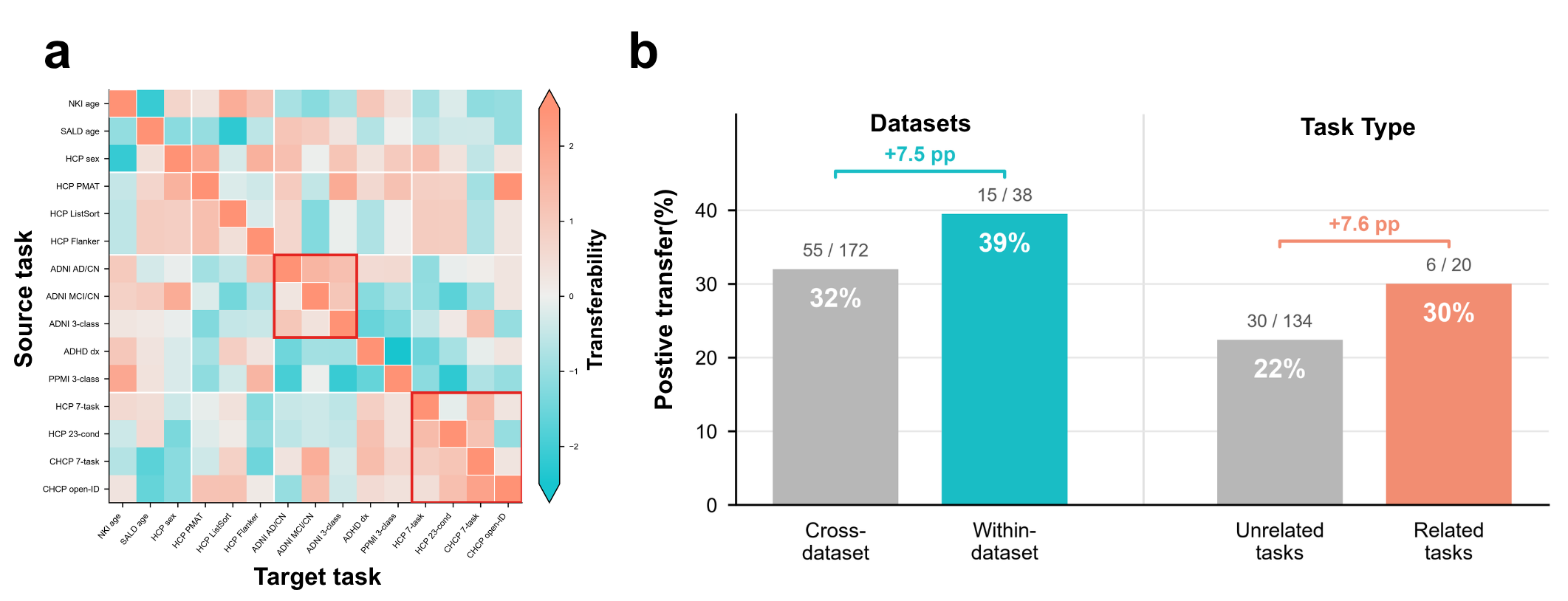}
    \caption{Downstream fMRI Taskonomy reveals directed transfer structure.
    (a) Column-wise z-scored first-order gains computed relative to the
    matched \(t\rightarrow t\) self-transfer reference. (b) Positive
    transfer is descriptively more prevalent within than across datasets,
    whereas the related-task contrast is modest.}
    \label{fig:downstream-transfer-structure}
\end{figure}

A direct route has \(S_r=\{t\}\) and \(u_{t,r}=0\), corresponding to the
self-transfer reference. Given a source budget \(K\) and maximum transfer order \(m\), we first
average route utility across targets within each of five predefined task
categories and then weight the categories equally:
\begin{equation}
\begin{aligned}
\max_{x,y}\quad&
\frac{1}{|\mathcal C|}
\sum_{c\in\mathcal C}
\frac{1}{|\mathcal T_c|}
\sum_{t\in\mathcal T_c}
\sum_{r\in\mathcal R_t^{(m)}}y_{t,r}u_{t,r}\\
\text{s.t.}\quad&
\sum_s x_s=K,\qquad
\sum_{r\in\mathcal R_t^{(m)}}y_{t,r}=1\quad\forall t,\\
&
y_{t,r}\leq x_s
\quad\forall t,r,\ s\in S_r,\qquad
x_s,y_{t,r}\in\{0,1\}.
\end{aligned}
\label{eq:bip}
\end{equation}
Here, \(\mathcal C\) denotes the five predefined task categories and
\(\mathcal T_c\) denotes the target tasks assigned to category \(c\). The solution selects exactly \(K\) supervised
sources and one route per target. Since route spaces are nested, a
maximum-order-four policy may still select an order-one or order-two
route.

\paragraph{Policy validation.}
BIP is solved only from construction-validation affinities, after which
the selected source portfolios and target routes are fixed. We then test
whether these policies generalize from the lightweight frozen
PCA--ridge construction protocol to a stronger target-adaptation
protocol: Brain-DiT remains frozen, while the routed source encoders and
a new target head are adapted on fresh low-shot support sets and
evaluated on sealed test subjects. We compare BIP-selected sources and
routes with BIP sources plus random feasible routes, fully random
policies, and capacity-matched scratch controls. A sequential
decomposition separates source-portfolio and route-selection effects.
Complete split, matching, aggregation, and uncertainty-estimation
procedures are provided in the supplementary material.

\begin{figure*}[!t]
    \centering
    \includegraphics[width=\textwidth]{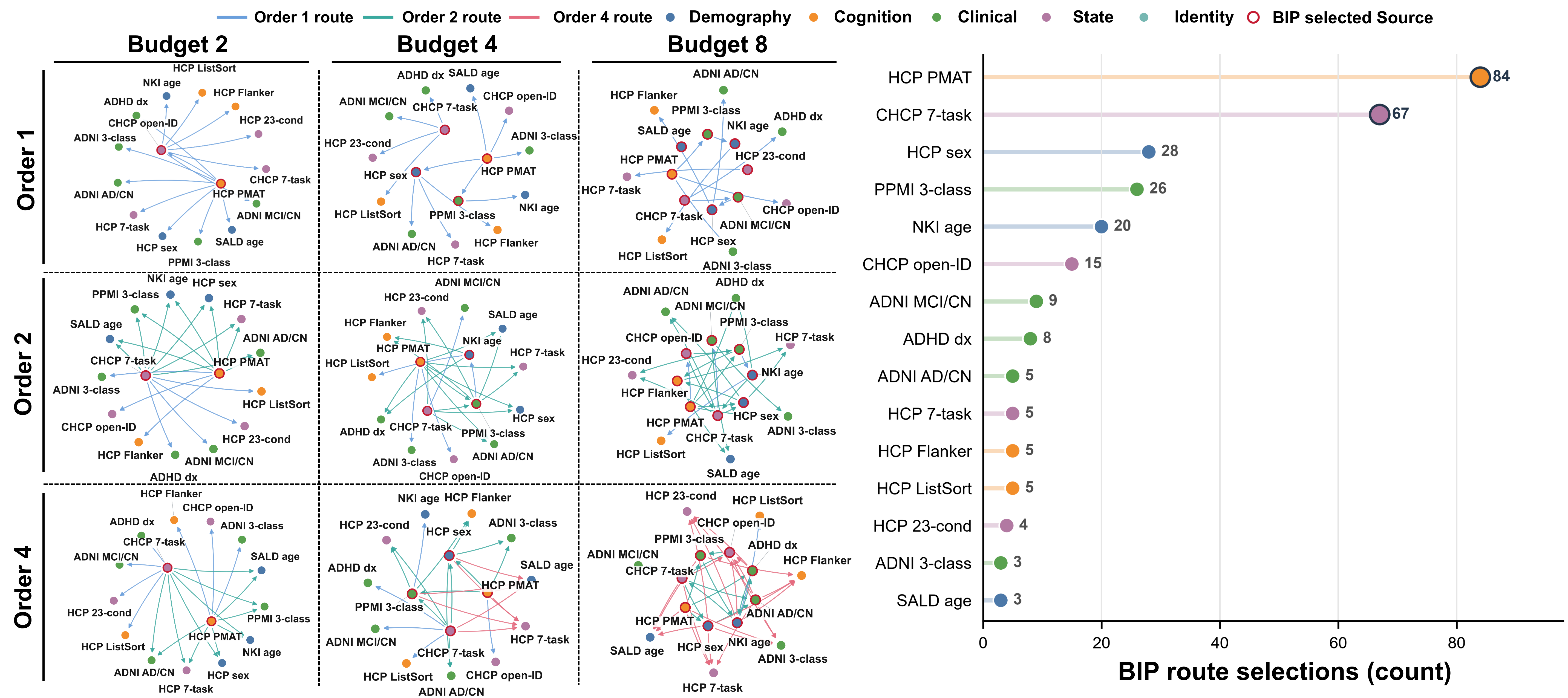}
    \caption{Budget-aware transfer policies derived from the downstream
    taskonomy. Left: BIP-selected directly supervised sources (red outlines)
    and target-specific routes across source budgets
    \(K\in\{2,4,8\}\) and maximum transfer orders
    \(m\in\{1,2,4\}\). Right: source-task usage counts aggregated over the
    frozen BIP policies, revealing recurrent transfer hubs.}
    \label{fig:budgeted-task-transfer}
\end{figure*}

\begin{figure}[!t]
    \centering
    \includegraphics[width=\columnwidth]{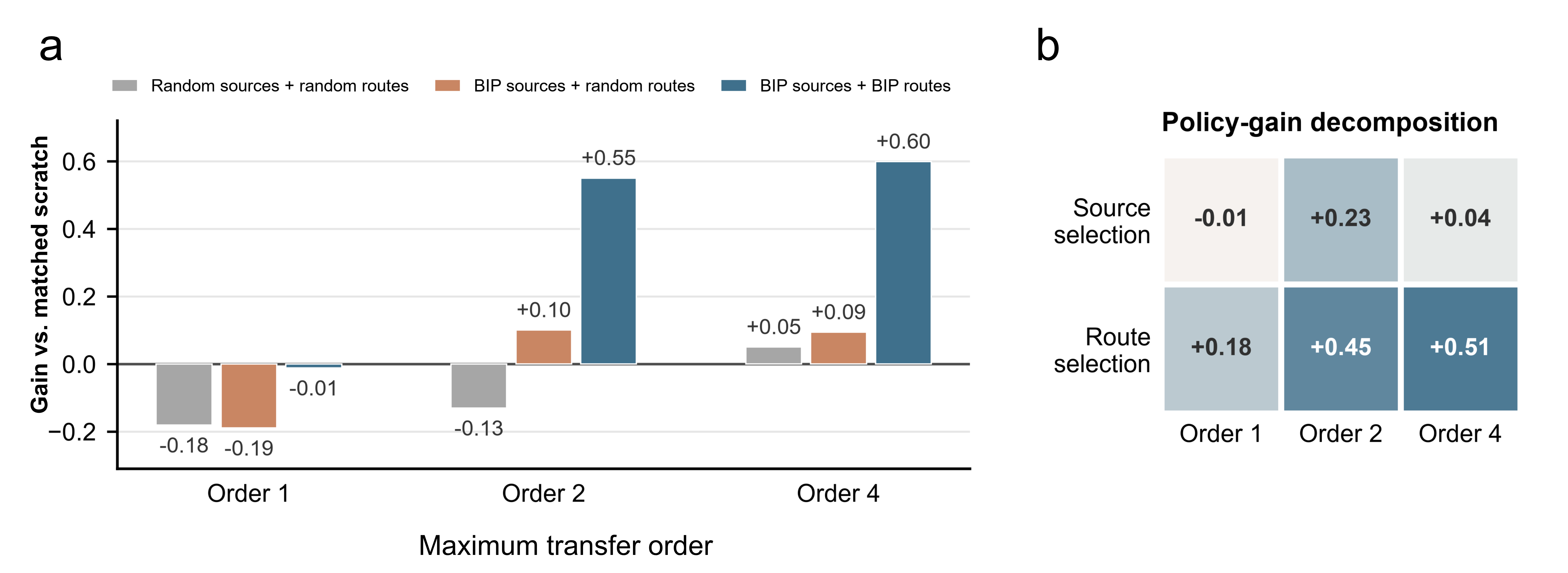}
    \caption{
    \textbf{Exploratory validation of Taskonomy-derived policies.}
    \textbf{a,} Gain over capacity-matched scratch for all 15 targets,
    first averaged within each of five predefined task categories and then
    equally across categories; values are further averaged over five
    support draws and budgets \(K\in\{2,4,8\}\).
    \textbf{b,} Decomposition of the BIP advantage into source- and
    route-selection contributions. Higher-order gains were driven mainly
    by route selection.}
    \label{fig:downstream-transfer-gain}
\end{figure}
\section{Experiments}
\label{sec:experiments}

\subsection{Experimental Setup}

\paragraph{Pretraining.}
We use ten fMRI domains spanning resting, task, naturalistic, lifespan,
and clinical data: HCP rest, HCP movie, HCP task, CHCP rest, ABCD rest,
NKI rest, ABIDE, ADHD, ADNI, and CineBrain
\cite{HCP,CHCP,ABCD,NKI,ABIDE,ADHD,ADNI,CineBrain}.
Following Omni-fMRI preprocessing~\cite{wang2026omni}, scans are mapped to
Schaefer-100 space and divided into non-overlapping 40-TR windows; splits
are subject-disjoint and normalization uses training data only.
SALD~\cite{SALD} is excluded from pretraining and held out for
out-of-domain age prediction. A lightweight unconditional Brain-DiT proxy estimates
domain difficulty and directed facilitation. We cross uniform and
high-to-low timestep schedules with uniform, random, and priority-guided
domain policies (Table~\ref{tab:curriculum-reconstruction}).
All policies share the same fixed-bank convergence criteria; their
update counts differ because each newly introduced domain or timestep
range changes the active training distribution.
Pretraining quality uses domain-balanced v-NMSE and 50-step DDIM reconstruction
PSD-NMSE and FC-MSE. Full preprocessing and training settings are provided
in the supplementary material.

\paragraph{Downstream evaluation.}
Following the Omni-fMRI benchmark~\cite{wang2026omni}, we evaluate full
fine-tuning on age regression for ABIDE and NKI and sex classification
for ABCD and HCP. We additionally evaluate held-out SALD age prediction
and held-out PPMI binary classification.
BrainLM, Brain-JEPA, and BrainMass serve as external foundation-model
baselines. Age prediction is evaluated using MSE and Pearson
correlation \(r\), while sex classification uses accuracy and macro-F1.
All Brain-DiT variants use identical data splits, prediction heads,
optimization settings, and three random seeds. Following Omni-fMRI, we
compare the best Brain-DiT result in each column with the strongest
external baseline using Cohen's \(d\), and mark large effects
(\(d\geq0.8\)) with an asterisk.

\paragraph{Taskonomy and policy validation.}
Participants are split once at the cohort level; all tasks from the same
cohort inherit the same mutually exclusive participant pools, and all
runs from one participant remain in a single pool. Using fixed features
from the official Brain-DiT checkpoint, we construct a 15-task taskonomy
through controlled first- and higher-order transfer and solve BIP using
only construction-validation affinities. The selected source portfolios
and target-specific routes are then frozen and evaluated on fresh low-shot
support sets and sealed test subjects against matched random-route, fully
random, and capacity-matched scratch controls. Full task definitions,
split statistics, adaptation settings, and matching protocols are
provided in the supplementary material.
\subsection{Domain Learning Structure Guides Curriculum Pretraining}

\paragraph{Domains differ in difficulty and directed facilitation.}
Figure~\ref{fig:domain-structure-curriculum}(a) shows a \(2.65\times\)
difference in proxy difficulty across the ten domains. NKI is the easiest
domain (\(D=0.245\)), followed by ABIDE and ADNI
(\(D=0.319\) and \(0.320\)), whereas HCP rest and HCP task are the most
difficult (\(D=0.639\) and \(0.650\)). This ranking does not simply
separate resting-state from task or naturalistic data, suggesting that
cohort, acquisition, and signal properties jointly affect learnability.

The compute-matched facilitation matrix in
Figure~\ref{fig:domain-structure-curriculum}(b) is strongly asymmetric.
Relative to generic non-target pretraining, HCP task yields \(4.1\%\) and
\(3.8\%\) directed facilitation for ADNI and NKI, whereas ADNI yields
little or negative relative benefit for most domains. Difficulty and
facilitation consequently produce an order that differs from a simple
easy-to-hard curriculum: ABCD is introduced first despite intermediate
difficulty, while the difficult HCP task domain is promoted by its strong
outgoing facilitation.

\paragraph{Domain and timestep curricula provide complementary benefits.}
Table~\ref{tab:curriculum-reconstruction} crosses two timestep policies
with three domain policies. Under uniform timesteps, Random and Priority
domain reduce v-NMSE to \(0.3950\) and \(0.3936\); under high-to-low
timesteps, Uniform and Random domain reach \(0.3885\) and \(0.3893\).
The high-to-low Priority row achieves the best v-NMSE
(\(0.3880\)), PSD-NMSE (\(0.1261\)), and FC-MSE (\(0.0547\)),
improving over uniform domains and timesteps by \(6.5\%\), \(16.3\%\),
and \(10.5\%\),
respectively.

The small gap between the high-to-low Uniform and Priority rows in v-NMSE
and FC-MSE indicates that
timestep scheduling explains most of the aggregate gain. Nevertheless,
the planned domain order provides a complementary benefit, most clearly
in PSD-NMSE (\(0.1302\rightarrow0.1261\)) and the downstream results
reported below.

\paragraph{Curriculum gains extend to in-domain tasks.}
As shown in Table~\ref{tab:downstream-comparison}, the high-to-low Priority
row is the strongest reported Brain-DiT configuration
on all four in-domain tasks. It achieves an MSE of \(0.43\) and
\(r=0.71\) on ABIDE age prediction, an MSE of \(0.27\) and \(r=0.87\)
on NKI age prediction, \(66.38\%\) accuracy and \(66.26\%\) macro-F1 on
ABCD sex classification, and \(84.32\%\) accuracy and \(84.29\%\)
macro-F1 on HCP sex classification. It also outperforms BrainLM,
Brain-JEPA, and BrainMass on these evaluations, indicating that the
curriculum benefit extends beyond the diffusion objective.

The held-out SALD results are less conclusive. Under high-to-low
timesteps, Random domain achieves the best MSE
(\(0.406\pm0.006\)), whereas Uniform domain achieves the best Pearson
correlation (\(0.756\pm0.002\)). Priority domain obtains
\(0.458\pm0.004\) and \(0.746\pm0.004\), respectively. Although all six
Brain-DiT configurations outperform the listed external baselines on
SALD, the split metric leadership indicates that no single domain policy
dominates this held-out task. In contrast, high-to-low Priority achieves
the best held-out PPMI results, with \(71.63\%\) accuracy and \(67.42\%\)
macro-F1. The contrasting SALD and PPMI results suggest that curriculum-induced representations transfer selectively rather than uniformly across unseen populations and clinical targets.

\subsection{Directed Task Relations Guide Budget-Aware Transfer}

\paragraph{Downstream transfer is asymmetric and target dependent.}
Figure~\ref{fig:downstream-transfer-structure}(a) visualizes first-order
transfer among the 15 downstream tasks. Although the heatmap is
standardized within each target for visualization, all analyses use the
original self-referenced transfer gains. A source that benefits one target
can be ineffective or harmful to another, and reversing the direction
generally changes the result, confirming that the taskonomy represents
directed transfer rather than symmetric similarity.

Figure~\ref{fig:downstream-transfer-structure}(b) further shows that
positive transfer occurs on \(39\%\) of within-dataset edges
(\(15/38\)), compared with \(32\%\) of cross-dataset edges
(\(55/172\)), a difference of \(7.5\) percentage points. Among pairs with
task-relatedness annotations, positive transfer occurs on \(30\%\) of
related edges (\(6/20\)) and \(22\%\) of unrelated edges
(\(30/134\)). These descriptive results suggest that both shared dataset
context and task semantics shape transfer, although neither factor alone
fully explains the observed structure.

\paragraph{BIP selects compact source portfolios and recurrent hubs.}
Figure~\ref{fig:budgeted-task-transfer} shows the policies selected across
budgets \(K\in\{2,4,8\}\) and maximum transfer orders
\(m\in\{1,2,4\}\). Under small budgets, BIP concentrates supervision on a
few sources and reuses them through multiple target-specific routes.
Larger budgets add more specialized sources and replace some indirect
routes with direct supervision. Allowing second- and fourth-order routes
further expands the set of targets supported by a fixed source portfolio.

The route-use analysis identifies several recurrent transfer hubs.
HCP PMAT and CHCP seven-task decoding are used most frequently, appearing
in 84 and 67 selected routes, respectively, followed by HCP sex, PPMI
three-class diagnosis, and NKI age. These frequencies are derived from
the frozen BIP policies and therefore reflect practical route utility,
rather than visual centrality in the first-order heatmap.

\paragraph{Exploratory policy validation favors higher-order planning.}
Figure~\ref{fig:downstream-transfer-gain} reports category-balanced gain over capacity-matched scratch across all 15 target tasks. BIP showed limited benefit under first-order transfer, whereas positive descriptive gains emerged when higher-order routes were available. Most of this improvement arose from target-specific route selection rather than source selection.

\section{Discussion and Conclusion}

We show that measured learning relations can organize fMRI foundation-model
pretraining and downstream adaptation without changing the backbone.
Random cumulative ordering improves over flat domain mixing, indicating
that staged exposure is itself beneficial. Priority-guided ordering adds
gains, most clearly in v-NMSE and spectral fidelity, while high-to-low-noise
scheduling accounts for most aggregate reconstruction improvement. Thus,
timestep and domain curricula are complementary: one organizes denoising
difficulty, whereas the other determines which experiences are introduced
first. Because policies plateau after different update counts, the
comparison reflects endpoint quality and convergence rather than
matched-compute efficiency.

Controlled downstream transfer reveals an asymmetric, target-dependent
taskonomy. BIP converts these relations into compact source portfolios and
target-specific routes, with descriptive gains when higher-order routes
are available. Sequential analysis associates these gains more strongly
with target-specific routing than source selection. Nevertheless, policy
validation remains exploratory, and task-dependent SALD and PPMI results
show that improved in-domain organization does not ensure uniform OOD
robustness. Future work should examine broader domain and task spaces,
cross-backbone stability, and cost-aware allocation of computation and
supervision.

\bibliography{bib/theory,bib/technique,bib/dataset}

\end{document}

%% file: authors.tex
\author{
    Junfeng Xia\textsuperscript{1}, Wenhao Ye\textsuperscript{1}, Junxiang Zhang\textsuperscript{1},\\
    Jiayu Zuo\textsuperscript{1}, Mo Wang\textsuperscript{2,1,4*}, Quanying Liu\textsuperscript{1,3,4*}
}
\affiliations{
    \textsuperscript{1}Department of Biomedical Engineering, Southern University of Science and\\
    Technology, Shenzhen, China\\
    \textsuperscript{2}Department of Computer Science, University of Warwick, The UK\\
    \textsuperscript{3}Shenzhen Loop Area Institute, Shenzhen, China\\
    \textsuperscript{4}Omni-Intelligence, Shenzhen, China\\
    \textsuperscript{*}Co-corresponding authors: 12250099@mail.sustech.edu.cn, liuqy@sustech.edu.cn
}